\documentclass[final]{cvpr}

\usepackage{times}
\usepackage{epsfig}
\usepackage{graphicx}
\usepackage{amsmath}
\usepackage{amssymb}
\usepackage{verbatim}
\usepackage[pagebackref=true,breaklinks=true,colorlinks,bookmarks=false]{hyperref}
\usepackage{url}
\usepackage{hyperref}

\begin{document}

%%%%%%%%% TITLE
\title{DeepSatData: Building large scale datasets of satellite images for training machine learning models}

\author{Michail Tarasiou\\
Imperial College London\\
% {\tt\small michail.tarasiou10@imperial.ac.uk}
% For a paper whose authors are all at the same institution,
% omit the following lines up until the closing ``}''.
% Additional authors and addresses can be added with ``\and'',
% just like the second author.
% To save space, use either the email address or home page, not both
\and
Stefanos Zafeiriou\\
Imperial College London\\
%First line of institution2 address\\
%{\tt\small secondauthor@i2.org}
}

\maketitle

%%%%%%%%% ABSTRACT
\begin{abstract}
   This report presents design considerations for automatically generating satellite imagery datasets for training machine learning models with emphasis placed on dense classification tasks, e.g. semantic segmentation. The implementation presented makes use of freely available {\it Sentinel-2} data which allows generation of large scale datasets required for training deep neural networks. We discuss issues faced from the point of view of deep neural network training and evaluation such as checking the quality of ground truth data and comment on the scalability of the approach. Accompanying code is provided in \href{https://github.com/michaeltrs/DeepSatData}{https://github.com/michaeltrs/DeepSatData}.   
\end{abstract}

%%%%%%%%% BODY TEXT
\section{Introduction}
Currently there are more than 150 satellites in orbit equipped with dedicated instruments gathering data for a variety of Earth Observation (EO) tasks. An ever increasing amount of that data are made freely accessible to the public, for example approximately $20Tb$ of new data are made available every day just through the European Space Agency's Sentinel 1-3 satellites.   \\
The {\it Copernicus Open Access Hub} (\href{https://scihub.copernicus.eu/}{COAH}) provides free and open access to data captured by the {\it European Space Agency's } \href{https://sentinel.esa.int/web/sentinel/missions}{Sentinel missions} starting from the In-Orbit Commissioning Review (IOCR). These data are made available directly through COAH either by use of a graphical user interface \cite{snap}, through a variety of platforms from the Copernicus Data and Information Access Services (DIAS) \cite{dias1, dias2, dias3, dias4, dias5} or through mirror sites \cite{mirror1, mirror2, mirror3, mirror4, mirror5, mirror6, mirror7, mirror8, mirror9, mirror10, mirror11}.\\
%provides tools for downloading Sentinel images (cite SNAP). In addition there exist a variety of APIs for accessing Sentinel data (cite). SEN4CAP can be cumbersome to install. \\
However, there are not, to the best of our knowledge, publicly available tools for downloading and processing Sentinel products at the scale required for successfully training machine learning models with satellite images. In this report we present {\it DeepSatData} a simple tool for downloading and processing Sentinel products from the point of view of training deep neural networks (DNN). With {\it DeepSatData} it is possible to automatically download available satellite imagery for a given area of interest (AOI) and time period of interest (POI) and to couple these with available ground truth data to create fully annotated datasets. In addition we present some general considerations for generating satellite imagery datasets suitable for training DNNs with particular emphasis on dense classification tasks, e.g semantic segmentation. 
%Rather than a fixed piece of work we intend to update this report with new features found in the  
%ng on the developpment of datasets for training machine learning models\\
%Question for GM: Can you make timeseries training data with SEN4CAP?
% provides complete, free and open access to Sentinel-1, Sentinel-2, Sentinel-3 and Sentinel-5P user products, starting from the In-Orbit Commissioning Review (IOCR)

\section{Background}

\subsection{Densely annotated data}
Typically, the anatomy of a dense classification dataset involves input arrays and dense annotations matching two or more dimensions of the inputs. In general obtaining annotations for dense classification tasks is a time consuming process. For example it is estimated that annotating a single image from the {\it Cityscapes} dataset \cite{cityscapes} fine set takes about 90min of work. For datasets where annotations are not included for all objects found in the inputs it is common practice to assign all multiple unknown objects into a single class which is either treated as an unknown or as part of a {\it background} class. Depending on the formulation of the task it is possible to treat the {\it background} class as another regular class or mask its influence during training and only learn to recognise the remaining classes.

\subsection{Dense classification tasks}
Similar to the general classification problem where the goal is to assign one of $N$ known classes to an input array, dense classification aspires to assign a class to every location, e.g. pixel, of an input array. Distinguishing between the different types of input arrays and the type of information encoded by the output classes can lead to defining several problems in computer vision. Inputs in general contain 2 or 3 spatial dimensions or a time dimension each with a fixed number of channels. For satellite imagery we are interested in either 2d images, i.e. a single image, or timeseries of images. In the second case each image is typically accompanied by a timestamp showing the capture time of the image. The interval between successive captures by a satellite is generally not constant. This is in contrast to video data, also consisting of timeseries of 2D images, in which there is a fixed time-step between successive frames. The model output most commonly encodes semantic or identity information or both leading to the tasks of {\it semantic segmentation} \cite{fcn, deeplab, deeplab2, deeplab3, psp, semseg1, semseg2, semseg3, dualattn}, {\it instance segmentation} \cite{maskrcnn, assemb, solo, seminst, ssap, pathagg, sgn} and joint semantic-instance segmentation \cite{panoptic, ubernet, jsis}.\\

\section{Downloading satellite data}
Downloading all required data for an AOI and POI can be a lengthy process. That is particularly the case for data captured more than 12 months in the past which will need to be accessed through the COAH's \href{https://scihub.copernicus.eu/userguide/LongTermArchive}{Long-Term Archive} (LTA). This means that the data will first have to be requested by the LTA and will be made available to download within 24h. Additionally, there is a maximum allowed number of requests per user to the LTA at a rate of 1 product request every 30min. In fact when working with annotated data it is most likely that these correspond to a period in the past thus all imagery products will need to be downloaded through the LTA. The limit in the amount of data that can be requested by the LTA poses a hard constraint on the number of products that can realistically be downloaded forcing us to optimize our selection process. Given the importance of selecting the right products in space and time we propose to spend some time manually selecting the products to download and automate the remaining part of the dataset generation process. Below are some general criteria for optimizing the product selection process.

\subsection{Low cloud cover ratio}
Cloud cover percentage is calculated for the full extent of a Sentinel product. While it can be the case that a clear image of the AOI can be found in a cloudy image (especially for small AOI) it is likely to get more clear images from products with low cloud coverage. Thus, we prioritize downloading the less cloudy images over the more cloudy ones. This parameter is controlled by the user defined variable "cloudcoverpercentage" in the start of each product selection script.

\subsection{Large overlap with the AOI}
Each Sentinel-2 tile covers a region of 100km x 100km which is large enough such that a single tile can be used for a dataset. For example using striding windows of 240m x 240m (24x24 pixels for the largest resolution band) results in approximately 200k samples. If the AOI is small it is quite likely that it will be covered by a single Sentinel tile in which case there is 100\% coverage of the AOI by that tile. If this is not the case then more than one products will need to be downloaded to cover the full extent of the AOI, in which case it is convenient to start with the ones that cover most of the AOI first. 

\subsection{Large product size}
As described in the \href{https://sentinel.esa.int/web/sentinel/user-guides/sentinel-2-msi/product-types}{S2 product description website} "Tiles can be fully or partially covered by image data. Partially covered tiles correspond to those at the edge of the swath.". In products partially covered with image data only part of the image contains information with remaining part covered by zero values. We prioritize downloading products with a small proportion of zero valued regions. 

\subsection{Uniformly spread along the time period of interest}
Modern Earth observation satellites can have a very small revisit time. For example the two satellites which form the Sentinel-2 constellation can have a revisit time of as few as 5 days. Rather than downloading products for all available dates during a POI we may need to subsample from available dates. Unless otherwise required by experimental settings we choose to select products such that they are spread as uniformly as possible during the POI.

\section{Data generation pipeline}

\begin{figure*}[!ht]
\centering
\includegraphics[width=\textwidth]{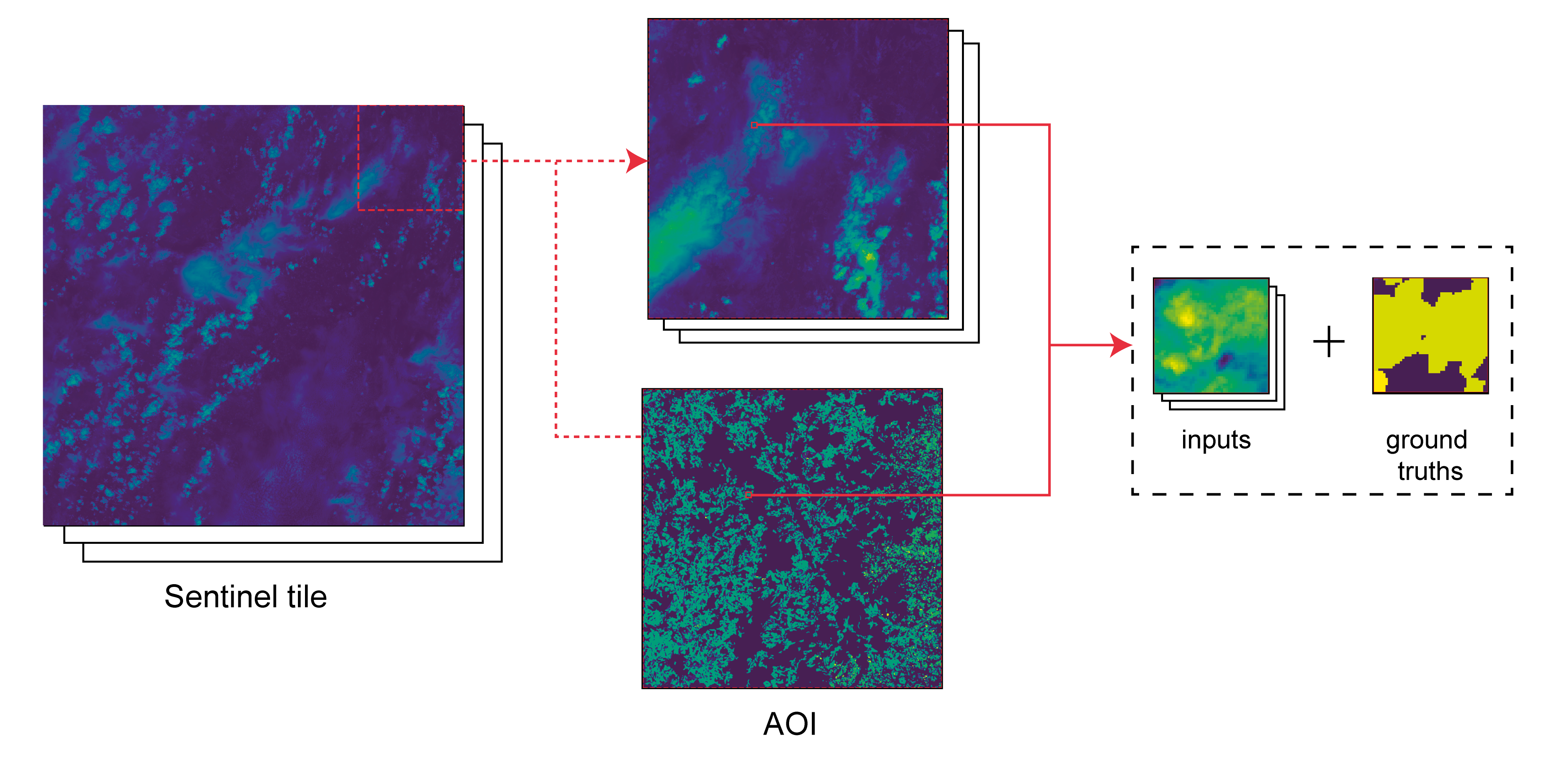}
\caption{Data generation process with ground truth annotations. Satellite product shown here (left) is a Sentinel-2 tile which covers a $100 \times 100$ $km^2$ area at a resolution of $10$ $m$. Extracted sample size (right) covers a $0.5 \times 0.5$ $km^2$ area.}
\label{process}
\end{figure*}

%zeroed out sections in S2 images, products
Having downloaded a set of satellite imagery products what is of interest is to extract small image patches of constant size that can fit into hardware accelerator memory and group/sort these patches by location into timeseries objects that can be used to train temporal models. Fig.\ref{process} shows this process and also indicates the relative size of typically extracted patches compared to the size of downloaded satellite products. Depending on whether there are available ground truth annotations we may choose to only process locations for which there are ground truths. These steps are further elaborated in the following sections. 
%make a timeseries of S2 images for the duration of one year. 
% he automated process for generating the dataset includes the following steps: defining an AOI and a specific time period of interest, extracting and rasterizing all relevant ground truth data, downloading all relevant satellite images, grouping and sorting satellite images to create timeseries objects and matching ground truths with timeseries by location. As a final step we create {\it boolean} masks for regions that should be masked out during training and inference. These include non-agricultural areas, crop types we do not wish to include as well as pixels contained to more than one parcels.
% If ground truth data are available we include a "labels" array (HxW) and an "ids" array (HxW) where the ground truth is stored for each location in the timeseries. The final form of a data sample used in the deep learning pipeline is a .pickle file containing one numpy array per S2 channel (not rescaled) and labels for a fixed size area (here 480mx480m), doy (day of year), year and location. he code assumes the following:

\subsection{From vector to raster ground truth data}

% \begin{figure*}[!tbp]
%   \centering
%   \begin{minipage}[b]{0.33\textwidth}
%     \includegraphics[width=\textwidth]{figures/parcels.png}
%     \caption{Parcel vector geometries}
%   \end{minipage}
%   \hfill
%   \begin{minipage}[b]{0.33\textwidth}
%     \includegraphics[width=\textwidth]{figures/parcels_labels.png}
%     \caption{Raster classes}
%   \end{minipage}
%   \hfill
%   \begin{minipage}[b]{0.33\textwidth}
%     \includegraphics[width=\textwidth]{figures/parcels_part_ass.png}
%     \caption{Doubly assigned pixels}
%   \end{minipage}
% \end{figure*}
\begin{figure*}[!tbp]
  \centering
  \includegraphics[width=0.9\textwidth]{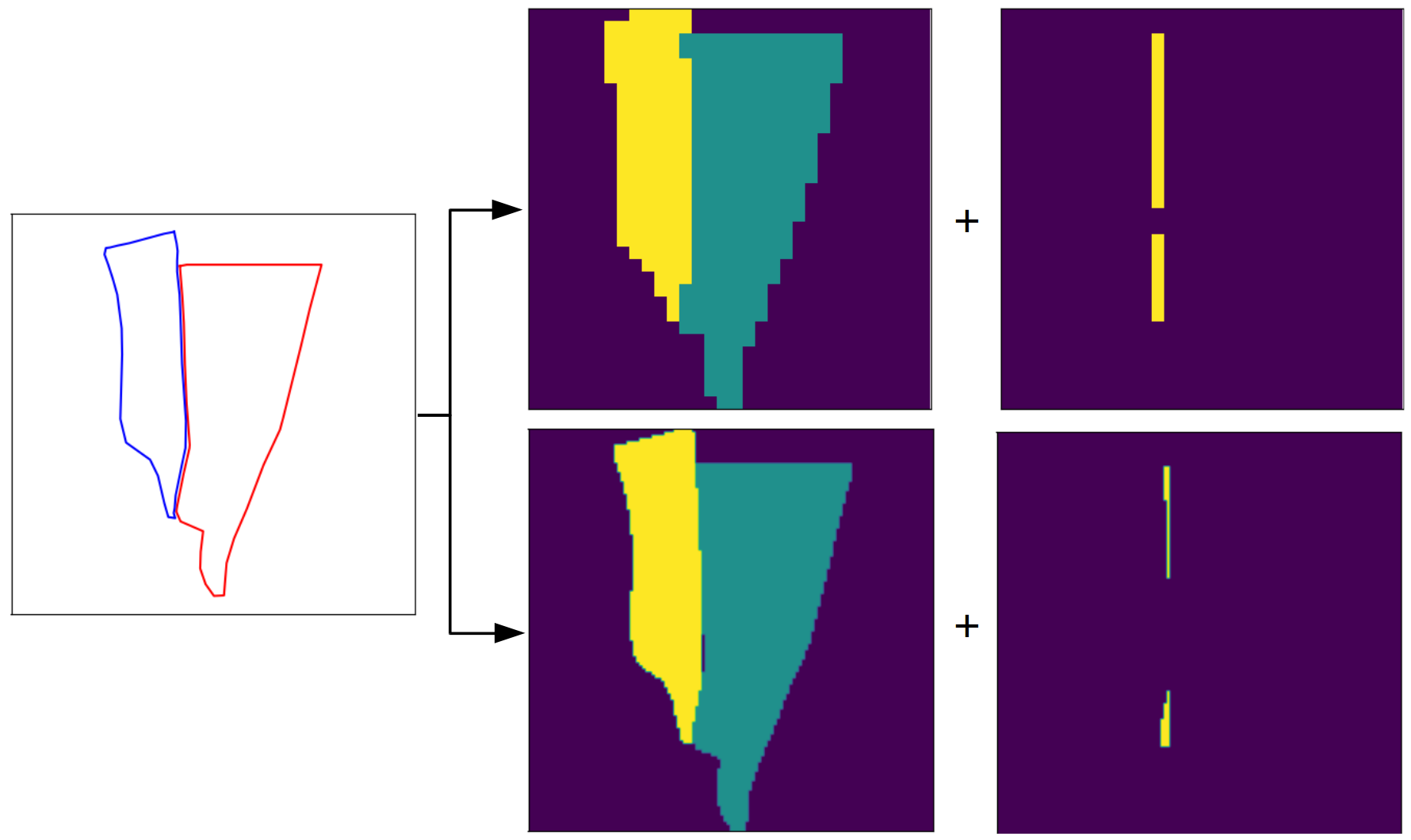}
  \caption{Parcel rasterization. We rasterize parcel vector geometries (left) to get a ground truth class map (center) and doubly assigned pixel masks (right). Two different resolutions shown at 10m (top) and 2.5m (bottom). WE note how the high resolution raster image contains a smaller ratio of doubly assigned pixels.}
  \label{raster}
\end{figure*}

\begin{figure*}[!tbp]
  \centering
  \includegraphics[width=0.9\textwidth]{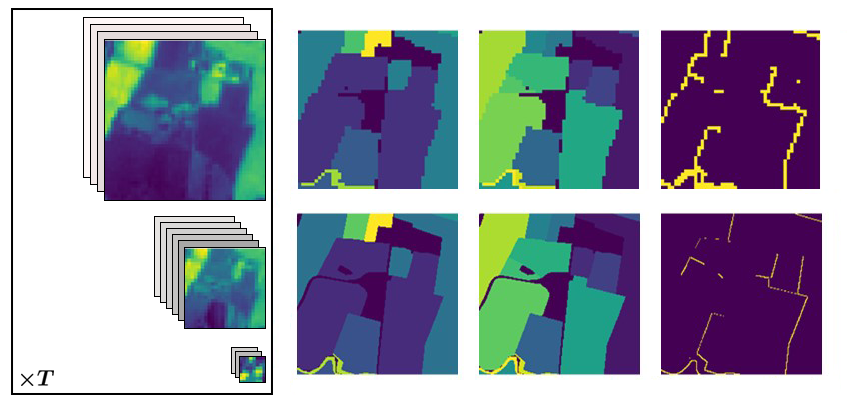}
  \caption{Example of data extracted using the {\it DeepSatData} pipeline. (left) Satellite image inputs. Here shown all four 10m bands, six 20m bands and three 60m bands for variable number of time steps {\it T} depending on location. (right) Ground truth data. From left to right we show crop types, parcel ids and doubly assigned pixels. Shown at 10m resolution (top row) and 2.5m $\times 4$ super-resolution (bottom row). If there are no available ground truths the pipeline extracts only the input data (left).}
  \label{sample}
\end{figure*}

This step is only relevant for cases when there are available ground truth annotations. We assume these collections are in the form of geo-polygons whose vertices are GPS coordinates at a given coordinate reference system (CRS) as this is the way typically agricultural ground truth data are collected. To ensure consistency we define a canonical form of representing such collections which includes the following fields for each agricultural parcel:
% show examples of polygons and generated raster images
\begin{itemize}
    \item \verb geometry  is a geo-polygon containing GPS coordinates for all the vertices of the agricultural parcel
    \item \verb crs  denotes the geographic CRS used 
    \item \verb ground_truth  indicates the class corresponding to area defined by \verb geometry. This is typically of type \verb int  for semantic or identity classes and typ \verb float  in the case of regression tasks
    \item \verb year  denotes the year the ground truth is valid for the given geometry
\end{itemize}
Using these data we follow a rasterization step. Here we first define a grid which is initiated by a value corresponding to a background class. For each pixel in the grid we calculate the ratio of the pixel area that is covered by the geo-polygon. All pixels partly or fully covered by the geo-polygon are assigned the \verb ground_truth  corresponding to that polygon. We note here that is is typical to define the grid size such that it equals the largest resolution satellite image available, however, this need not necessarily be the case. Using CNNs it is straightforward to control the output resolution of our model to match the rasterization resolution. Also, for crop-type semantic segmentation \cite{cscl} showed that it is possible to successfully learn to distinguish crop types at a higher resolution than satellite pixels. An example of performed rasterization is shown in Fig.\ref{raster}.

\subsubsection{Masking ground truth inconsistencies}
% Types of errors, satellite image resolution 10m, similar to GPS errors. 
The process of generating dense ground truth annotations for geodata is unique w.r.t other dense labelled data, e.g natural images \cite{cityscapes, coco, voc}, in that source images and ground truths are first collected separately and are then aligned by geolocation. While a human annotator working on a semantic segmentation dataset will draw semantic classes on top of captured images, ground truth collection for remote sensing involves a step of gathering GPS coordinates on the field and a separate step of matching these with source images. This introduces the possibility for systematic geolocation errors, the gathered GPS coordinates might not be in complete agreement with the geolocation corresponding to the satellite images. While it is possible to identify some cases where there are noticeable offsets between inputs and ground truths by inspection, \cite{mtlcc} identify single pixel offset errors, in general it is impossible for a human to correct all such mistakes. 
For this reason we are using boolean masks to mark inconsistencies when it is possible to identify, such as pixels that fall inside multiple polygons during the rasterization step. We distinguish between pixels that are partly or fully claimed by two or more polygons. While the former case is a natural outcome of the rasterization step and is improved with using a higher resolution grid, as shown in Fig.\ref{raster} low vs high resolution, the latter case clearly indicates a geocoding error in either polygon.

\subsection{Splitting a Sentinel product to small windows and making timeseries objects}
% memory considerations
The main reason for choosing to split a satellite imagery product into smaller, equal size patches is the requirement to load multiple timeseries objects into hardware accelerator memory for efficient training. While the size of a satellite image is in the order of tens or hundreds of $km$, e.g a single Sentinel-2 tile covers a $100 \times 100$ $km^2$ area, sizes typically used for semantic segmentation are in the order of hundreds of $m$, e.g $240m$ \cite{mtlcc}, $480m$ \cite{mtlcc, cscl}, $640m$ \cite{Rustowicz2019SemanticSO}, an example of that scale difference can be seen in Fig.\ref{process}. We may choose to split the AOI only for locations where ground-truth annotations are available or, as is the case for unsupervised learning tasks, we may choose to split the entire AOI. 

% \subsection{Making timeseries objects}
The end result of the data generation process is a set of time-series objects containing image patches corresponding to the same location at different timestamps. Even though it would be possible to load all satellite images for all timestamps in memory and split-save to disk in one step this can be forbidding in terms of memory consumption for a long POI. For this reason we choose to separate the steps of extracting patches and grouping/sorting these in time to create the final outputs. An example of data included in a single sample point extracted using the {\it DeepSatData} pipeline can be seen in Fig.\ref{sample}. 

% \subsection{Data splits for training, validation and testing}
\section{Conclusion}
This report presented {\it DeepSatData} a pipeline for automatically generating data for training machine learning models on earth observation tasks and explained the main considerations behind its design.    
While particular emphasis was placed on generating datasets for dense classification tasks using time-series of satellite images it is trivial to extend the provided code to extract single images for dense or global classification. We intend to provide such capabilities in future updates.

{\small
\bibliographystyle{ieee_fullname}
\bibliography{ms}

\begin{thebibliography}{10}\itemsep=-1pt

\bibitem{dias1}
\url{https://creodias.eu/}.

\bibitem{dias2}
\url{https://sobloo.eu/}.

\bibitem{dias3}
\url{https://mundiwebservices.com/}.

\bibitem{dias4}
\url{https://www.wekeo.eu/}.

\bibitem{dias5}
\url{https://www.onda-dias.eu/cms/}.

\bibitem{mirror1}
\url{https://sentinels.space.noa.gr/}.

\bibitem{mirror2}
\url{https://copernicus.nci.org.au/sara.client/\#/home}.

\bibitem{mirror3}
\url{https://finhub.nsdc.fmi.fi/\#/home}.

\bibitem{mirror4}
\url{https://data.sentinel.zamg.ac.at/dhus/\#/home}.

\bibitem{mirror5}
\url{https://peps.cnes.fr/rocket/\#/home}.

\bibitem{mirror6}
\url{https://code-de.org/de/}.

\bibitem{mirror7}
\url{https://www.collgs.lu/}.

\bibitem{mirror8}
\url{https://colhub.met.no/\#/home}.

\bibitem{mirror9}
\url{https://ipsentinel.ipma.pt/dhus/\#/home}.

\bibitem{mirror10}
\url{https://sedas.satapps.org/}.

\bibitem{mirror11}
\url{http://swea.rymdstyrelsen.se/portal/}.

\bibitem{snap}
{SNAP-ESA} sentinel application platform v2.0.2.
\newblock \url{http://step.esa.int}.

\bibitem{seminst}
Bert Brabandere, Davy Neven, and Luc Van~Gool.
\newblock Semantic instance segmentation with a discriminative loss function.
\newblock 08 2017.

\bibitem{deeplab}
L. {Chen}, G. {Papandreou}, I. {Kokkinos}, K. {Murphy}, and A.~L. {Yuille}.
\newblock Deeplab: Semantic image segmentation with deep convolutional nets,
  atrous convolution, and fully connected crfs.
\newblock {\em IEEE Transactions on Pattern Analysis and Machine Intelligence},
  40(4):834--848, 2018.

\bibitem{deeplab2}
Liang{-}Chieh Chen, George Papandreou, Florian Schroff, and Hartwig Adam.
\newblock Rethinking atrous convolution for semantic image segmentation.
\newblock {\em CoRR}, abs/1706.05587, 2017.

\bibitem{deeplab3}
Liang{-}Chieh Chen, Yukun Zhu, George Papandreou, Florian Schroff, and Hartwig
  Adam.
\newblock Encoder-decoder with atrous separable convolution for semantic image
  segmentation.
\newblock {\em CoRR}, abs/1802.02611, 2018.

\bibitem{cityscapes}
Marius Cordts, Mohamed Omran, Sebastian Ramos, Timo Rehfeld, Markus Enzweiler,
  Rodrigo Benenson, Uwe Franke, Stefan Roth, and Bernt Schiele.
\newblock The cityscapes dataset for semantic urban scene understanding.
\newblock {\em CoRR}, abs/1604.01685, 2016.

\bibitem{semseg3}
H. {Ding}, X. {Jiang}, B. {Shuai}, A.~Q. {Liu}, and G. {Wang}.
\newblock Context contrasted feature and gated multi-scale aggregation for
  scene segmentation.
\newblock In {\em 2018 IEEE/CVF Conference on Computer Vision and Pattern
  Recognition}, pages 2393--2402, 2018.

\bibitem{voc}
M. Everingham, S.~M.~A. Eslami, L. Van~Gool, C.~K.~I. Williams, J. Winn, and A.
  Zisserman.
\newblock The pascal visual object classes challenge: A retrospective.
\newblock {\em International Journal of Computer Vision}, 111(1):98--136, Jan.
  2015.

\bibitem{dualattn}
Jun Fu, Jing Liu, Haijie Tian, Yong Li, Yongjun Bao, Zhiwei Fang, and Hanqing
  Lu.
\newblock Dual attention network for scene segmentation.
\newblock In {\em Proceedings of the IEEE/CVF Conference on Computer Vision and
  Pattern Recognition (CVPR)}, June 2019.

\bibitem{semseg2}
Jun Fu, Jing Liu, Yuhang Wang, and Hanqing Lu.
\newblock Stacked deconvolutional network for semantic segmentation.
\newblock {\em CoRR}, abs/1708.04943, 2017.

\bibitem{ssap}
Naiyu Gao, Yanhu Shan, Yupei Wang, Xin Zhao, Yinan Yu, Ming Yang, and Kaiqi
  Huang.
\newblock {SSAP:} single-shot instance segmentation with affinity pyramid.
\newblock {\em CoRR}, abs/1909.01616, 2019.

\bibitem{maskrcnn}
K. {He}, G. {Gkioxari}, P. {Dollár}, and R. {Girshick}.
\newblock Mask r-cnn.
\newblock In {\em 2017 IEEE International Conference on Computer Vision
  (ICCV)}, pages 2980--2988, 2017.

\bibitem{panoptic}
Alexander Kirillov, Kaiming He, Ross Girshick, Carsten Rother, and Piotr
  Dollar.
\newblock Panoptic segmentation.
\newblock In {\em Proceedings of the IEEE/CVF Conference on Computer Vision and
  Pattern Recognition (CVPR)}, June 2019.

\bibitem{ubernet}
I. {Kokkinos}.
\newblock Ubernet: Training a universal convolutional neural network for low-,
  mid-, and high-level vision using diverse datasets and limited memory.
\newblock In {\em 2017 IEEE Conference on Computer Vision and Pattern
  Recognition (CVPR)}, pages 5454--5463, 2017.

\bibitem{semseg1}
Di Lin, Yuanfeng Ji, Dani Lischinski, Daniel Cohen-Or, and Hui Huang.
\newblock Multi-scale context intertwining for semantic segmentation.
\newblock In {\em Proceedings of the European Conference on Computer Vision
  (ECCV)}, September 2018.

\bibitem{coco}
Tsung{-}Yi Lin, Michael Maire, Serge~J. Belongie, Lubomir~D. Bourdev, Ross~B.
  Girshick, James Hays, Pietro Perona, Deva Ramanan, Piotr Doll{\'{a}}r, and
  C.~Lawrence Zitnick.
\newblock Microsoft {COCO:} common objects in context.
\newblock {\em CoRR}, abs/1405.0312, 2014.

\bibitem{sgn}
S. {Liu}, J. {Jia}, S. {Fidler}, and R. {Urtasun}.
\newblock Sgn: Sequential grouping networks for instance segmentation.
\newblock In {\em 2017 IEEE International Conference on Computer Vision
  (ICCV)}, pages 3516--3524, 2017.

\bibitem{pathagg}
S. {Liu}, L. {Qi}, H. {Qin}, J. {Shi}, and J. {Jia}.
\newblock Path aggregation network for instance segmentation.
\newblock In {\em 2018 IEEE/CVF Conference on Computer Vision and Pattern
  Recognition}, pages 8759--8768, 2018.

\bibitem{fcn}
Jonathan Long, Evan Shelhamer, and Trevor Darrell.
\newblock Fully convolutional networks for semantic segmentation.
\newblock {\em CoRR}, abs/1411.4038, 2014.

\bibitem{assemb}
Alejandro Newell, Zhiao Huang, and Jia Deng.
\newblock Associative embedding: End-to-end learning for joint detection and
  grouping.
\newblock In I. Guyon, U.~V. Luxburg, S. Bengio, H. Wallach, R. Fergus, S.
  Vishwanathan, and R. Garnett, editors, {\em Advances in Neural Information
  Processing Systems}, volume~30. Curran Associates, Inc., 2017.

\bibitem{jsis}
Quang{-}Hieu Pham, Duc~Thanh Nguyen, Binh{-}Son Hua, Gemma Roig, and Sai{-}Kit
  Yeung.
\newblock {JSIS3D:} joint semantic-instance segmentation of 3d point clouds
  with multi-task pointwise networks and multi-value conditional random fields.
\newblock {\em CoRR}, abs/1904.00699, 2019.

\bibitem{Rustowicz2019SemanticSO}
Rose Rustowicz, Robin Cheong, Lijing Wang, Stefano Ermon, Marshall Burke, and
  David~B. Lobell.
\newblock Semantic segmentation of crop type in africa: A novel dataset and
  analysis of deep learning methods.
\newblock In {\em CVPR Workshops}, 2019.

\bibitem{mtlcc}
Marc Rußwurm and Marco Körner.
\newblock Multi-temporal land cover classification with sequential recurrent
  encoders.
\newblock {\em ISPRS International Journal of Geo-Information}, 7(4):129, Mar
  2018.

\bibitem{cscl}
Michail Tarasiou, Riza~Alp G{\"{u}}ler, and Stefanos Zafeiriou.
\newblock Context-self contrastive pretraining for crop type semantic
  segmentation.
\newblock {\em CoRR}, abs/2104.04310, 2021.

\bibitem{solo}
Xinlong Wang, Tao Kong, Chunhua Shen, Yuning Jiang, and Lei Li.
\newblock {SOLO:} segmenting objects by locations.
\newblock {\em CoRR}, abs/1912.04488, 2019.

\bibitem{psp}
H. {Zhao}, J. {Shi}, X. {Qi}, X. {Wang}, and J. {Jia}.
\newblock Pyramid scene parsing network.
\newblock In {\em 2017 IEEE Conference on Computer Vision and Pattern
  Recognition (CVPR)}, pages 6230--6239, 2017.

\end{thebibliography}
}

\end{document}